\pdfoutput=1

\documentclass[11pt]{article}

\usepackage[preprint]{acl}
\usepackage{listings}
\usepackage{xcolor}
\usepackage{tcolorbox}
\usepackage{multicol}
\lstdefinestyle{mystyle}{
    backgroundcolor=\color{white},
    commentstyle=\color{green!50!black},
    keywordstyle=\color{blue},
    numberstyle=\tiny\color{gray},
    stringstyle=\color{red}, 
    basicstyle=\ttfamily\footnotesize,
    breakatwhitespace=false,
    breaklines=true,
    captionpos=b,
    keepspaces=true,
    numbers=left,
    numbersep=5pt,
    showspaces=false,
    showstringspaces=false,
    showtabs=false,
    tabsize=2,
    frame=single,
    rulecolor=\color{black},
    xleftmargin=10pt,
    xrightmargin=10pt,
}
\usepackage{amssymb}
\usepackage{ulem}
\usepackage{times}
\usepackage{latexsym}
\usepackage[ruled,vlined]{algorithm2e}
 \SetAlFnt{\small}
\usepackage[T1]{fontenc}
\usepackage{CJKutf8}

\usepackage[utf8]{inputenc}

\usepackage{microtype}
\usepackage{booktabs}
\usepackage{arydshln}
\usepackage{inconsolata}
\usepackage{multirow} 
\usepackage{graphicx}

\usepackage{soul}
\usepackage{color}
\usepackage{xcolor}
 \usepackage{amsmath} 
\usepackage{colortbl}
\usepackage{float}
\definecolor{softyellow}{rgb}{0.98, 0.98, 0.82}

\definecolor{example}{RGB}{95, 157, 241} 
\sethlcolor{example}

\newcommand{\modelname}{{\fontfamily{ppl}\selectfont KnowCoder-\textbf{X}}\xspace}

\newcommand{\datasetname}{{\fontfamily{ppl}\selectfont ParallelNER}\xspace}

\newcommand{\class}[1]{\texttt{#1}}
\newcommand{\variable}[1]{\textit{#1}}
\newcommand{\fun}[1]{\emph{#1}}

\newcommand{\green}[1]{\textcolor{green!40!black}{\texttt{#1}}}

\newcommand{\sref}[1]{\S\ref{#1}}

\title{\modelname: Boosting Multilingual Information Extraction via Code}

\author{
 \textbf{Yuxin Zuo\textsuperscript{1,2,3}}\thanks{\ \ Co-first authors.},
 \textbf{Wenxuan Jiang\textsuperscript{4}}\footnotemark[1],
 \textbf{Wenxuan Liu\textsuperscript{1,2,3}}\footnotemark[1],
 \textbf{Zixuan Li\textsuperscript{1,2}}\thanks{\ \ Corresponding authors.},
 \textbf{Long Bai\textsuperscript{1,2}},
\\ \textbf{Hanbin Wang\textsuperscript{5}},
 \textbf{Yutao Zeng\textsuperscript{1}},
 \textbf{Xiaolong Jin\textsuperscript{1,2,3}}\footnotemark[2],
 \textbf{Jiafeng Guo\textsuperscript{1,2,3}},
 \textbf{Xueqi Cheng\textsuperscript{1,2,3}}
\\
 \textsuperscript{1}Key Laboratory of Network Data Science and Technology, \\Institute of Computing Technology, Chinese Academy of Sciences
 \\\textsuperscript{2}State Key Laboratory of AI Safety
 \\\textsuperscript{3}School of Computer Science, University of Chinese Academy of Sciences
 \\\textsuperscript{4}School of Software, Northeastern University
 \\\textsuperscript{5}School of Software, Peking University
\\
 \texttt{\{zuoyuxin24s, liuwenxuan2024z, lizixuan24h, jinxiaolong\}@ict.ac.cn}
}

\begin{document}

\begin{CJK}{UTF8}{gkai}

\maketitle
\begin{abstract}
Empirical evidence indicates that LLMs exhibit spontaneous cross-lingual alignment.
However, although LLMs show promising cross-lingual alignment in Information Extraction (IE), a significant imbalance across languages persists, highlighting an underlying deficiency.
To address this, we propose \modelname, a powerful code LLM with advanced cross-lingual and multilingual capabilities for universal IE.
Firstly, it standardizes the representation of multilingual schemas using Python classes, ensuring a consistent ontology across different languages. Then, IE across languages is formulated as a unified code generation task.
Secondly, we conduct IE cross-lingual alignment instruction tuning on the translated instance prediction task to enhance the model's cross-lingual transferability.
During this phase, we also construct a high-quality and diverse bilingual IE parallel dataset with $\mathbf{257k}$ samples, called \textbf{\datasetname}, synthesized by our proposed robust three-stage pipeline, with manual annotation to ensure quality.
Although without training in $\mathbf{29}$ unseen languages, \modelname surpasses ChatGPT by $\mathbf{30.17\%}$ and SoTA by $\mathbf{20.03\%}$, thereby demonstrating superior cross-lingual IE capabilities.
Comprehensive evaluations on $\mathbf{64}$ IE benchmarks in Chinese and English under various settings demonstrate that \modelname significantly enhances cross-lingual IE transfer through boosting the IE alignment.
Our code and dataset are available at: \url{https://github.com/ICT-GoKnow/KnowCoder}.


\end{abstract}

\normalem
\section{Introduction}

Cross-lingual Information Extraction~(IE) focuses on automatically extracting structured information from texts in unseen languages, following manually designed schemas.
Cross-lingual schema understanding and representation have long been a challenge.
Large Language Models~(LLMs), trained on massive multilingual corpora, have significantly advanced multilingual language processing~\citep{brown2020language,touvron2023llama2}.
Previous studies~\citep{qi-etal-2023-cross,wang2024seaeval,gao2024multilingual,li2024prealign} have shown that LLMs exhibit spontaneous cross-lingual alignment to facilitate the transfer of abilities and knowledge across languages.
Our findings suggest the presence of this alignment in IE, indicating a strong potential for improving the IE cross-lingual transfer.
We first define IE parallel data, which is across various languages that share the same schema, sentences, and extracted instances.
Previous studies used label projection to generate such data to alleviate the challenges of IE in low-resource~\cite{kolluru2022alignment,hennig2023multitacred,rios2024transalign}.
We construct an IE parallel dataset in Chinese based on CoNLL 2003~\cite{conll_2003} manually, to evaluate the LLaMA2-7B trained on English Named Entity Recognition~(NER) datasets used in KnowCoder~\cite{li2024knowcoder}.
Table~\ref{tab:intro} shows the results.
After training on English IE datasets, there was a notable enhancement in the parallel dataset of Chinese IE~($5.9$$\rightarrow$$52.7$), strongly supporting the presence of IE cross-lingual alignment.
However, the significant performance gaps between the two languages~($95.1$ \textit{vs} $52.7$) indicate that the alignment remains weak.
To address this, we propose two strategies to enhance IE cross-lingual alignment in LLMs by aligning the schema and extraction process across different languages, respectively:

\begin{table}[!tbp]
\centering
\resizebox{\columnwidth}{!}{
\begin{tabular}{@{}lcc@{}}
\toprule
Model & CoNLL 2003 & CoNLL 2003~(zh) \\ \midrule
LLaMA2-7B & $26.9$     &  $5.9$    \\
LLaMA2-7B (w/ training) & $95.1$     & $52.7$    \\
\bottomrule
\end{tabular}
}
\caption{Results on CoNLL 2003 and CoNLL 2003~(zh). The latter is the constructed Chinese parallel dataset.}
\vspace{-4mm}
\label{tab:intro}
\end{table}

Firstly, we standardize IE schemas across languages, particularly non-English ones, into unified Python classes.
X-GEAR~\cite{huang2022multilingual} introduced a language-agnostic output template to maintain multilingual IE uniformity, aiding cross-lingual transfer.
However, this uniformity is not thorough due to the absence of a schema in the input context, which hinders both cross-lingual transfer and multilingual schema understanding.
Moreover, its highly customized input and output limit scalability to UIE and LLM-based IE.
Thus, we introduce a unified UIE schema representation in the input and output, ensuring thorough uniformity in multilingual IE to improve cross-lingual alignment within LLMs.
We also observe that object-oriented features in code are well-suited for unified schema representation and knowledge sharing across languages.
Although code-based approaches~\cite{wang2022code4struct,li2024knowcoder,liu2025towards} have shown promising results in IE, their effectiveness in cross-lingual IE still needs to be validated.
Therefore, we leverage Python classes to represent multilingual schemas, ensuring consistent ontology representation across languages and facilitating efficient IE knowledge transfer.
The consistent schema allows the model to learn and efficiently share the knowledge of the same ontology across languages.

\begin{figure}[tbp]
    \centering
    \includegraphics[width=\linewidth]{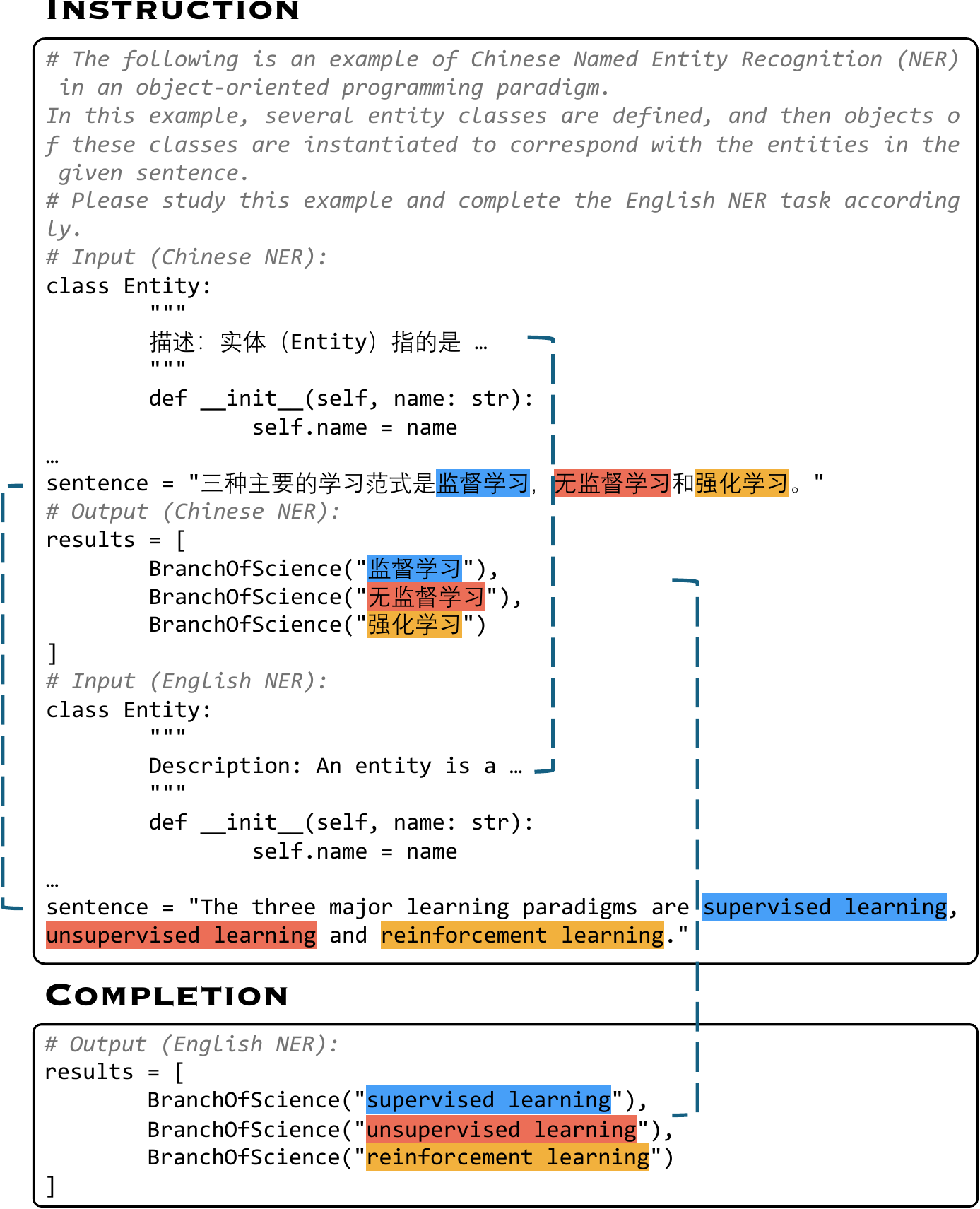}
    \caption{
    An example of instruction-tuning data used in the IE cross-lingual alignment phase.
    The \hl{highlighted span} in the completion and the two same color \hl{highlighted spans} in the instruction must ensure semantic consistency and be exactly matched, respectively.}
    \vspace{-4mm}
    \label{fig:align_data_case}
\end{figure}

Secondly, we introduce an IE cross-lingual alignment instruction tuning phase.
We first propose a task that predicts target language instances, aiding in the alignment of the extraction process~(see Figure~\ref{fig:align_data_case} for an example).
Using prediction tasks with parallel data to enhance multilingual alignment through instruction tuning is a widely adopted paradigm~\citep{zhu2023extrapolating,gao2024multilingual}.
The key challenge is ensuring that extracted instances maintain semantic consistency with their source-language counterparts while appearing in the target language sentence.
Unlike directly predicting complete IE parallel data, which focuses on aligning sentences, our method prioritizes the alignment of translated instances, which is the central goal of IE cross-lingual alignment.

To use high-quality IE parallel data for the phase, we propose an automatic three-stage LLM-based pipeline, which achieves $99\%$ average faithfulness across $10$ languages evaluated on the label projection benchmark WikiANN~\cite{wikiann}.
Utilizing the pipeline, we construct a high-quality and diverse NER English-Chinese parallel dataset \datasetname, using the GPT-4o series model supplemented with manual annotation.
By distilling the IE alignment capabilities of advanced proprietary models, the IE cross-lingual alignment phase enhances cross-lingual generalization in IE tasks.

Ultimately, we obtain the \modelname through Chinese and English IE instruction tuning.
We evaluate the \modelname on $64$ benchmarks for the NER, Relation Extraction~(RE), Event Detection~(ED), and Event Argument Extraction~(EAE) tasks in Chinese and English.
\modelname exhibits superior cross-lingual generalization across $29$ unseen diverse languages, and surpasses ChatGPT by $30.17\%$ and SoTA by $20.03\%$.
Moreover, it achieved an impressive average improvement of $11.43\%$ over the SoTA across $20$ low-resource African languages.
In the supervised evaluation, \modelname consistently ranks within the top-$2$ results across $40$ (of $42$) benchmarks.
Notably, \modelname achieves SoTA across all Chinese IE benchmarks, effectively demonstrating the strength of knowledge transfer from English IE through our method.
Our contributions can be summarized as:
\begin{itemize}
    \item The code-based multilingual IE method of \modelname unifies the representation of schemas across different languages, thereby boosting cross-lingual transfer.
    \item We incorporate an IE cross-lingual alignment phase to improve cross-lingual transfer. This phase involves instruction tuning on a newly proposed translated instance prediction task.
    \item We propose a robust three-stage IE parallel data construction pipeline and construct a high-quality bilingual parallel NER dataset \datasetname with $257,190$ samples to provide valuable resources to the community.
\end{itemize}

\section{Related Work}
\noindent \textbf{Transfer Learning for cross-lingual IE}  \ \ 
\citet{huang2017zero} introduced a zero-shot learning method for EE, facilitating generalization across unseen languages. CLaP~\cite{huang2022multilingual} utilized generative language models for cross-lingual EAE. Prompt-XRE~\cite{hsu2023prompt} enhanced cross-lingual RE through prompt-based learning, reducing reliance on large annotated datasets.
\citet{zubillaga2024event} emphasized the significance of typology in cross-lingual EE, particularly for low-resource languages like Basque.
In contrast, \modelname focuses on leveraging the spontaneous alignment of LLMs to facilitate cross-lingual IE transfer on all IE tasks.

\noindent \textbf{LLMs for Multilingual IE} \ \ 
YAYI-UIE~\cite{xiao2023yayi} proposes a chat-enhanced instruction tuning framework for UIE.
IEPILE~\cite{gui-etal-2024-iepile} collects a comprehensive bilingual
IE instruction corpus, and B\textsuperscript{2}NER~\cite{yang2024beyond} further designs a universal entity taxonomy. However, previous works overlook the mutual influence mechanisms between schemas in different languages and lack a unified representation.

\noindent \textbf{Code-Based IE} \ \ 
Code-based IE aims to formulate the IE as a code generation task.
CodeIE~\cite{codeie} first leverages the Code-LLM and recasts IE tasks into Python function completion tasks.
Code4Struct~\cite{wang2022code4struct} represents the schema in Python classes for the EAE task.
Code4UIE~\cite{guo2023retrieval} further constructs code-based schema for all the UIE tasks.
Besides, KnowCoder~\cite{li2024knowcoder} utilizes class inheritance, class methods, and type hints to further enrich the schema representation.
GoLLIE~\cite{sainz2024gollie} incorporates guidelines within class comments.
However, these works focus on the English IE task, and we are dedicated to exploring the significant role of code in multilingual IE.

\section{Code-Based Multilingual IE} \label{sec:code-based-IE}

\modelname formulates the IE tasks across various languages using a unified code generation framework.
In this section, we introduce the instruction and completion of the proposed code-based multilingual IE.
The instruction consists of two parts~(\sref{sec:instruction-format}):
(1) schema representation code across languages, including class definition and comments;
(2) task prompts for completion.
Subsequently, we present the completion format~(\sref{sec:completion-format}).

\subsection{Instruction Format}\label{sec:instruction-format}

\textbf{Class Definition} \ \ 
The IE schema comprises multiple concepts, each of which consists of a name and a set of attributes.
The three fundamental concepts of IE schemas [\class{Entity}, \class{Relation}, \class{Event}] are first defined as base classes.
For each concept of a schema, a corresponding class is subsequently defined.
Each concept inherits from its respective base class by default.
Then, we define the attributes of each concept (such as argument roles) as input arguments of the constructor~\fun{\_\_init\_\_()}.
Specially, for IE in non-English languages, we first create a mapping from the original schema to the English schema.
The mapping standardizes the representation of the same ontology across different languages by utilizing the unified Python class.
For example, the entity concept \begin{CJK}{UTF8}{mj}\class{\green{사람}}\end{CJK} in Korean and \begin{CJK}{UTF8}{gkai}\class{\green{人物}}\end{CJK} in Chinese are both mapped to the \class{class \textbf{\green{PER(Entity)}}}, ensuring consistent semantic representation across linguistic boundaries to align the schema in different languages.

\noindent \textbf{Class Comments} \ \ 
We utilize class comments to provide clear definitions of concepts similar to GoLLIE~\cite{sainz2024gollie} and KnowCoder~\cite{li2024knowcoder} for the understanding of concepts in different languages.
Class comments comprise two components:
1) examples contain instances corresponding to the concept.
2) a description that explains the concept;
For the examples part, we sample up to $10$ instances for each concept by frequency.
For the description part, we follow the annotation guidelines of datasets by default.
We propose an LLM-based approach to generate descriptions for datasets where guidelines are unavailable.
To enable the LLM to better understand the concepts and the annotation style of the dataset, we first sample $10$ instances to prompt the LLM to summarize an initial description.
To avoid providing excessive examples at once to hinder effective induction summarization, we then sample $20$ instances for the LLM to iteratively refine the description based on each one.
Specifically, for each instance, if it cannot be categorized into the corresponding concept according to the description, the LLM will adjust the inappropriate parts of the current description accordingly.
We use the \texttt{GPT-4o-2024-08-06} as the LLM.
Appendix~\ref{appx:description-generation} provides details on the description generation procedure.

\noindent \textbf{Task Prompt} \ \ 
The task prompt includes a docstring defining the task and a string variable \variable{sentence} that holds the text. The prompt is then concatenated with the schema to form the instruction.

\subsection{Completion Format}\label{sec:completion-format}
The completion represents the final output generated by the model, which begins after the assignment operation (i.e., \variable{results =}).
Each completion is a list, where each element is an instantiated object, representing the extracted structured knowledge instance.
Additionally, we utilize the \fun{exec()} function in Python to execute the instructions and completions of all code-based data in this work to ensure quality.
We also use this function to execute the generated code to extract the results.

\section{IE Cross-Lingual Alignment}
\modelname adopts a two-phase instruction tuning framework.
The first phase is the IE cross-lingual alignment phase, where we propose the task and data.
Subsequently, we conduct instruction tuning on Chinese and English UIE tasks with $46$ IE datasets to obtain \modelname.
Appendix~\ref{appx:examples} shows examples of instruction tuning data.
Appendix~\ref{appx:data-statics} shows the detailed dataset statistics.
In this section, we introduce the IE cross-lingual alignment phase.
First, we introduce the instruction and completion of the translated spans prediction task~(\sref{sec:trans_task}).
To obtain high-quality IE parallel data for this training phase, we propose a three-stage automatic pipeline for constructing IE parallel data~(\sref{sec:pipeline}) and creating \datasetname~(\sref{sec:parallelner}).

\subsection{Translated Instances Prediction Task}\label{sec:trans_task}

The translated instances prediction task is the training objective of this phase, aiming to perform IE in the target language.
The instruction starts with a task description, prompting the LLM to predict translated instances by concatenating the input-output pairs of source-language IE data with the input of target-language IE data to form the instruction.
The completion contains the output of the target language data.
Figure~\ref{fig:align_data_case} shows an example.

\normalem
\begin{algorithm}[t]
	\caption{Sample Translation}
	\label{alg:algorithm1}
       
	\SetKwInOut{Initialize}{Initialize}
	\SetKwInOut{Output}{Output}
	\SetKwFunction{TemplateVerify}{TemplateVerify}
	\SetKwFunction{TemplateComlete}{TemplateComlete}

        \SetKwFunction{TemplateT}{TemplateT}
        \SetKwFunction{TemplateI}{TemplateI}
        \SetKwFunction{TemplateS}{TemplateS}
        \SetKwFunction{Concat}{Concat}

        \SetKwFunction{LLM}{LLM}
         \SetKwFunction{JointTranslate}{JointTranslate}
         \SetKwFunction{SpanRephrase}{SpanRephrase}
         \SetKwFunction{SentenceRephrase}{SentenceRephrase}
         \SetKwFunction{Main}{Main}
         \SetKwProg{Fn}{Function}{:}{}
        \Initialize{Source Sentence $s^{src}$ \\  Source Spans $I^{src}$}
	\Output{Target Sentence $s^{tgt}$ \\  
    Target Spans $I^{tgt}$}
    \SetKwFunction{Translate}{Translate}
    \Fn{\JointTranslate{$s^{src}, I^{src}$}}{
        $ prompt = $ \TemplateT($s^{src}, I^{src}$)  \; 
        \textit{// Obtain Prompt for Joint Translation.}\\
        $I^{tgt}, s^{tgt} = \LLM(prompt) $\; 
    \textbf{return}  $I^{tgt}, s^{tgt}$   \;
    }
    \Fn{\SpanRephrase{$I^{tgt}, I^{src}, s^{tgt}$}}{
        $Flag = False $ \;
        \ForEach{$(i^{tgt}_{m}, i^{src}_{m})$ in $(I^{tgt}, I^{src})$}{
            \If{$i^{tgt}_{m}$ not in $s^{tgt}$}{
                $ prompt = $ 
                \TemplateI($s^{tgt}, i^{src}_{m}, i^{tgt}_{m}$)\; 
                
                \textit{// Obtain Prompt for Span Rephrase.}
                $ i^{cor}_{m} = $ \LLM($prompt$) \;

          \eIf{$i^{cor}_{m}$ in $s^{tgt}$}{
                   $ I^{tgt}.update(i^{tgt}_{m}, i^{cor}_{m} ) $ \;
                } { $Flag = $ True \;}
            }
        } 
    \textbf{return} $I^{tgt}, Flag$   \;  }
    \Fn{\SentenceRephrase{$I^{tgt}, s^{src}$}}{
            $ prompt = $ \TemplateS($I^{tgt}, s^{src}$) ;
            
            \textit{// Obtain Prompt for Sentence Rephrase.}
            
            $ s^{tgt} = $ \LLM($prompt$); 
    \textbf{return} $s^{tgt}$  \;
    }
    \SetKwFunction{SentenceRephrase}{SentenceRephrase}
    \SetKwProg{Fn}{Function}{:}{}
    \Fn{\Main{$s^{src}, I^{src}$}}{
        $I^{tgt}, s^{tgt} = $  \JointTranslate{$s^{src}, I^{src}$};\\
        $I^{tgt}, Flag = $  \SpanRephrase{$I^{tgt}, I^{src}, s^{tgt}$};\\
        \If{Flag == True}{
            $s^{tgt} = $  \SentenceRephrase{$I^{tgt}, s^{src}$}; \\
        }
        \textbf{return}   $s^{tgt}, I^{tgt}$
        
    }
    
\label{alg}
\end{algorithm}

\subsection{IE Parallel Data Construction Pipeline}\label{sec:pipeline}
The construction of IE parallel data involves two key components: sample translation~(i.e., label projection) and schema translation.
The sample consists of sentences and spans, where spans refer to the annotated segments within the sentence that are relevant to the IE task~\cite{kripke1971identity, chen2004detecting}.
Given the sentence $s^{src}$ and span list $I^{src} = \{ i^{src}_{1}, i^{src}_{2}, ..., i^{src}_{n} \}$ in the source language, label projection aims to obtain the sentence $s^{tgt}$ and span list $I^{tgt} = \{i^{tgt}_{1},i^{tgt}_{2}, ..., i^{tgt}_{n}\}$ in the target language.
Schema translation can be carried out through machine translation and manual annotation.
The primary challenge lies in the sample translation, which involves two key issues~\cite{parekh-etal-2024-contextual}:
1) \textbf{Inaccurate Translation}: Either the spans or the sentences are inaccurately translated;
2) \textbf{Missing Spans}: The translated spans may be missing from the translated sentence.

\newpage
Traditional label projection methods~\cite{dou-neubig-2021-word, chen-etal-2023-frustratingly, parekh-etal-2024-contextual}, such as CLaP, typically first generate $s^{tgt}$, and then obtain $I^{tgt}$ through label retrieval, word alignment, or contextual translation.
However, these methods have a critical flaw: errors introduced during the generation of $s^{tgt}$ are propagated into the process of obtaining $I^{tgt}$.
For example, during the translation of $s^{src}$ to get $s^{tgt}$, $i^{src}_{k}$ may be mistranslated or merged, especially when $i^{src}_{k}$ is a pronoun such as ``his'', leading to the failure in obtaining the correct $i^{tgt}_{k}$.
To address the issue, we propose a three-stage pipeline that implements joint translation and multi-grained rephrasing. Appendix~\ref{appx:prompts} shows all prompts used in the pipeline.

\noindent \textbf{Stage 1: Joint Translation} \ \
We propose the first stage to ensure semantic consistency between the translated sentence $s^{tgt}$ and the span list $I^{tgt}$.
Compared to previous methods, where $s^{tgt}$ and $I^{tgt}$ are obtained sequentially, this stage utilizes LLMs to simultaneously translate and generate both $s^{tgt}$ and $I^{tgt}$, thereby reducing error accumulation through \textit{parallel processing}.
This stage provides $s^{src}$ and $I^{src}$ as a comprehensive context that introduces inner constraints on the two key issues, thus enabling a more accurate and natural translation.
Specifically, we use a predefined template to format the sentence $s^{src}$ along with spans $I^{src}$ to serve as a prompt for the LLM to perform joint translation.

\noindent \textbf{Stage 2: Span Rephrase} \ \ 
Stage 2 and 3 tackle the missing spans issue by rephrasing.
In Stage 2, $i^{tgt}_{k}$ corresponding to $i^{src}_{k}$ is identified from $s^{tgt}$, similar to the CLaP~\cite{parekh-etal-2024-contextual}, which allows the rephrasing of the previously translated ones to further address the missing spans issue that persists after Stage 1.
Specifically, if $i^{tgt}_{m} \notin s^{tgt}$, we prompt the LLM to find the span $i^{tgt-cor}_{m}$ in $s^{tgt}$ that semantically matches with $i^{src}_{m}$.
If $i^{tgt-cor}_{m} \in s^{tgt} $, $i^{tgt}_{m}$ will be rephrased as $i^{tgt-cor}_{m}$.

\noindent \textbf{Stage 3: Sentence Rephrase} \ \ 
However, due to the inherent issue of error accumulation in the traditional \textit{sentence-then-span} approach employed by CLaP-like methods, we introduce Stage 3.
In this stage, we rephrase the translated sentence that is a \textit{span-then-sentence} strategy, thereby mitigating the error propagation of the \textit{sentence-then-span} approach and addressing the missing spans issue that remains unresolved.
Specifically, if not all $i^{tgt}_{m} \in s^{tgt}$, we prompt the LLM to rephrase $s^{tgt}$ based on $s^{src}$ to ensure all $i^{tgt}_{m}$ are included.

This pipeline is capable of generating IE parallel data across any two languages, as demonstrated in Algorithm~\ref{alg}.
By integrating three translation strategies, \textit{parallel processing}, \textit{sentence-then-span}, and \textit{span-then-sentence}, we have achieved unprecedented performance on the label projection task, approaching an accuracy rate of nearly 100\%.
Following~\citet{parekh-etal-2024-contextual}, we evaluate the pipeline on the label projection benchmark WikiANN~\cite{wikiann}, a multilingual IE dataset, and achieve an average $99\%$ on the faithfulness evaluation in $10$ English-centric bilingual settings.
Appendix~\ref{app:pipeline-evaluation} shows the detailed settings and results.

\subsection{\datasetname}\label{sec:parallelner}

Since alignment training requires including bilingual IE samples within a single prompt, the excessive concepts in RE and EE result in insufficient context length of LLMs.
Thus, we construct NER parallel data for the phase.
NER serves as a shared and foundational task in the IE framework, enabling entity identification through spotting abilities to support all IE tasks~\cite{lu-etal-2022-unified}.

Using the pipeline, we construct a high-quality English~(en) $\Leftrightarrow$ Chinese~(zh) NER parallel dataset, \datasetname.
To ensure linguistic diversity, we use high-quality datasets of each language: WikiNeural~\cite{tedeschi2021wikineural} (en$\Rightarrow$zh) and CLUENER2020~\cite{xu2020cluener2020} (zh$\Rightarrow$en), resulting $\mathbf{257,190}$ samples.
We use the \texttt{GPT-4o-mini} as the base LLM of the pipeline.
If the process fails at any stage, such as occurring Missing Spans, we employ \texttt{GPT-4o-2024-08-06} to facilitate reprocessing.
With our pipeline, the missing spans issue is rarely observed, with a rate of $15/10000$
in CLUENER2020 and $82/92720$~($8.84$\textpertenthousand) in WikiNeural.
For the $97$ samples above, we conduct further manual annotation to ensure the quality.
We utilize \datasetname to construct the training data for this phase in both language directions~(en$\Leftrightarrow$zh) to facilitate cross-lingual transfer.
Appendix~\ref{app:parallel} shows the statistics of \datasetname.

\begin{table*}[]
\centering
\resizebox{\textwidth}{!}{
\begin{tabular}{@{}l|cc|ccccccccc|cc@{}}
\toprule
\multirow{2}{*}{Model} &  &  & \multicolumn{8}{c}{\textit{\quad Cross-Lingual}} &  & \multirow{2}{*}{\textbf{\textit{Avg$_{cross}$}}} & \multirow{2}{*}{\textbf{\textit{Avg}}} \\
 & English & Chinese & German & Spanish & Dutch & Russian & Bengali & Persian & Hindi & Korean & Turkish &  &    \\ \midrule
ChatGPT$^{\dag}$   & 37.20   & 18.80   & 37.10  & 34.70   & 35.70 & 27.40   & 23.30   & 25.90   & 27.30 & 30.00  & 31.90   & 30.37      & 29.94 \\

YAYI-UIE~\cite{xiao2023yayi}
&52.04 	&37.89 	&41.78 	&42.44 	&38.02 	&30.97 	&15.97 	&21.96 	&20.93 	&26.45 	&30.74 	&29.92 	&32.65 \\

IEPILE~\cite{gui-etal-2024-iepile}    &53.19 	&39.26 &	39.87 &	42.41 &	35.69 &	28.69 &	12.26 &	23.93 &	18.98 &	27.56 &	26.93& 	28.48 &	31.71 \\

GLiNER~\cite{zaratiana-etal-2024-gliner}     & 41.70   & 24.30$^{\ast}$   & 39.50  & 42.10   & 38.90 & 33.30   & \textbf{25.90}   & 30.20   & 27.80 & 28.70  & 30.00   & 32.93      & 32.95 \\

B\textsuperscript{2}NER~\cite{yang2024beyond} & 54.80   & 45.40   & 36.60  & 46.00   & 43.00 & 33.90   & -       & -       & -     & -      & -       & -          & -     \\
\midrule
\modelname &
\textbf{56.37} &
\textbf{47.53} &
\textbf{49.87} &
\textbf{54.17} &
\textbf{48.81} &
\textbf{41.53} &
24.13 &
\textbf{30.83} &
\textbf{29.67} &
\textbf{33.56} &
\textbf{43.19} &
\textbf{39.53$^{\uparrow\textbf{20.03\%}}$} &
\textbf{41.79} \\
\rowcolor{example}
Supervised~\cite{Malmasi2022MultiCoNERAL} & 62.70   & 53.10   & 64.60  & 58.70   & 62.60 & 59.70   & 39.70   & 52.30   & 47.80 & 55.80  & 46.80   & 54.22      & 54.89 \\ \bottomrule
\end{tabular}
}
\caption{Results of cross-lingual evaluation in Multiconer22. \textbf{\textit{Avg$_{cross}$}} represents the average performance across $9$ unseen languages. $^{\textbf{\dag}}$ indicates that the score is reported by~\citet{Lai2023ChatGPTBE}.
$^{\textbf{\ddag}}$ indicates that the score is from our implementation.
$^{\ast}$ indicates that the result of GLiNER in Chinese is under the cross-lingual setting.
\hl{Supervised} denotes the supervised method, which is a strong baseline since the above methods are under the zero-shot setting.} 
\label{tab:cross-ood}
\vspace{-4mm}
\end{table*}

\section{Experimental Settings}

\subsection{Implementation Details}

We use span-based offset Micro-F1 to evaluate the methods.
For NER, an entity is considered correct if the span and type match a golden annotation.
For RE, a relation is considered correct if 
its type, subject entity, and object entity 
match a golden annotation.
For ED, an event is valid if its trigger and type match a golden annotation.
For EAE, given event type and trigger, an argument is valid if the span and its role match a golden annotation.

\modelname is fine-tuned based on Baichuan2-7B-Base~\cite{baichuan2023baichuan2} using LLaMA-Factory~\cite{zheng2024llamafactory}.
We apply LoRA~\cite{hulora} with a LoRA rank of $32$ for parameter-efficient fine-tuning.
The warmup ratio is set to $0.01$, and the learning rate for both phases is $3 \times 10^{-4}$.
The sequence length is limited to $4096$, and the batch size is $256$.
During inference, the temperature is set to $0$.
All experiments are conducted on 8 x NVIDIA A8000 80G GPUs.

\subsection{Datasets}
We follow the IEPILE~\cite{gui-etal-2024-iepile}, YAYI-UIE~\cite{xiao2023yayi}, and B\textsuperscript{2}NER~\cite{yang2024beyond} to conduct a comprehensive evaluation.
Overall, for English IE, we evaluate performance across $28$ benchmarks for NER, $10$ benchmarks for RE, $6$ benchmarks for ED, and $3$ benchmarks for EAE.
For Chinese IE, we evaluate performance on $4$ benchmarks for NER, $5$ benchmarks for RE, $4$ benchmarks for ED, and $2$ benchmarks for EAE.

Specifically, for cross-lingual evaluation, we evaluate performance on the multilingual NER benchmark Multiconer22~\cite{malmasi-etal-2022-multiconer} and the low-resource African language NER benchmark MasakhaNER 2.0~\cite{adelani2022masakhaner}.
Moreover, we conduct the supervised evaluation on various benchmarks: $23$ NER, $8$ RE, $3$ ED, and $3$ EAE benchmarks in English, as well as $2$ NER, $2$ RE, $2$ ED, and $2$ EAE benchmarks in Chinese.
Appendix~\ref{appx:data-statics} presents the detailed statistics and usage.

We only train on the supervised evaluation IE datasets of the three main baselines to ensure a fair comparison, without additional training data.

\subsection{Baselines}\label{sec:baselines}

To enable a fair and comprehensive comparison, we mainly use two types of baselines:
Following B\textsuperscript{2}NER~\cite{yang2024beyond}, we first compare \modelname with BERT-Base~\cite{devlin2018bert}.
Additionally, we compare \modelname with the SoTA systems for English and Chinese IE based on LLMs, including IEPILE~\cite{gui-etal-2024-iepile}, YAYI-UIE~\cite{xiao2023yayi}, and B\textsuperscript{2}NER~\cite{yang2024beyond}, which are our main baselines and based on natural language prompts.
The three baselines exhibit varying degrees of unfairness compared to \modelname.
IEPILE and YAYI-UIE are fine-tuned on the larger backbone Baichuan2-13B, while B\textsuperscript{2}NER utilizes the more advanced backbone Intern2LM-7B ~\cite{cai2024internlm2}.

For the cross-lingual evaluation, we further compare \modelname with ChatGPT~(from \citealp{Lai2023ChatGPTBE}) and GLiNER~\cite{zaratiana-etal-2024-gliner}.

\section{Experimental Results}
Our extensive evaluation of IE cross-lingual transfer can be divided into two main dimensions: the \textit{cross-lingual evaluation} for external-language transfer in unseen languages~(\sref{sec:cross-lingual-evaluation}), and the \textit{supervised evaluation} for internal-language transfer between English and Chinese~(\sref{sec:supervised-evaluation}).
We additionally conduct the zero-shot evaluation for internal language transfer.
Appendix~\ref{appx:zero-shot} shows the results.

\subsection{Cross-Lingual Evaluation}\label{sec:cross-lingual-evaluation}
To evaluate whether \modelname improves IE cross-lingual alignment, we conduct the cross-lingual evaluation on $9$ unseen languages of Multiconer22.
We have also supplemented the evaluation in Chinese and English.
Table~\ref{tab:cross-ood} shows results.
\modelname significantly outperforms ChatGPT and SoTA in the cross-lingual setting, achieving a $30.17\%$ and $20.03\%$ improvement, respectively.
For $4$ unseen languages used in B\textsuperscript{2}NER that constitute more than $0.1\%$ of the general LLM pretraining corpus~\cite{touvron2023llama2}, \modelname outperforms B\textsuperscript{2}NER by $19.36\%$, with results comparable to B\textsuperscript{2}NER in both Chinese and English, demonstrating a significant improvement in cross-lingual transfer.
Surprisingly, despite no training data in $9$ unseen languages, \modelname has achieved performance comparable to the supervised method in Spanish and Turkish that significantly outperforms the SoTA.

Appendix \ref{appx:cross-ood-low} presents the evaluation results on MasakhaNER 2.0 with $20$ African low-resource languages.
The results demonstrate that \modelname exhibits strong cross-lingual performance and can extend to a wide range of languages, particularly those with limited resources.

\begin{table}[!tbp]
\centering
\resizebox{1.0\columnwidth}{!}{
\begin{tabular}{@{}lccccc@{}}
\toprule
Dataset & BERT & YAYI-UIE & IEPILE & B\textsuperscript{2}NER & \modelname \\ \midrule
ACE 2005 & \uline{87.30} & 81.78 & 81.86 & 83.04 & \textbf{87.49} \\
AnatEM & 85.82 & 76.54 & 87.21 & \uline{89.18} & \textbf{89.19} \\
BC2GM & 80.90 & \uline{82.05} & 80.73 & 81.95 & \textbf{84.49} \\
BC4CHEMD & 86.72 & 88.46 & \textbf{90.45} & 88.96 & \uline{89.57} \\
BC5CDR & 85.28 & 83.67 & 88.07 & \textbf{88.52} & \uline{88.46} \\
Broad Twitter	&58.61&	\textbf{83.52}&	-	&82.16&	\uline{82.36} \\
CoNLL 2003 & 92.40 & \textbf{96.77} & 92.49 & 92.56 & \uline{94.69} \\
FabNER & 64.20 & 72.63 & 77.07 & \uline{78.82} & \textbf{83.19} \\
FindVehicle & 87.13 & 98.47 & \uline{98.49} & 97.89 & \textbf{99.47} \\
GENIA & 73.30 & 75.21 & \uline{76.66} & 76.43 & \textbf{78.97} \\
HarveyNER & \textbf{82.26} & 69.57 & 67.70 & 73.67 & \uline{73.91} \\
MIT Movie & 88.78 & 70.14 & 88.23 & \textbf{90.78} & \uline{89.50} \\
MIT Restaurant & 81.02 & 79.38 & 79.85 & \textbf{83.71} & \uline{81.95} \\
MultiNERD & 91.25 & 88.42 & \uline{94.60} & 93.98 & \textbf{95.94} \\
NCBI & 80.20 & \textbf{87.29} & 85.26 & 84.83 & \uline{85.49} \\

OntoNotes &	\textbf{91.11}&	87.04&	87.55	&84.31	& \uline{87.91}\\
PolyglotNER	& \textbf{75.65}&	\uline{70.85}& 	-&	61.96 &	64.47 \\
TweetNER7&	56.49 &	\uline{66.99} 	&-	&66.26 	& \textbf{67.98} \\
WikiANN	&70.60 &	72.63&	-	& \textbf{85.07}&	\uline{84.69} \\
wikiNeural	&82.78	&87.63	&-&	\textbf{93.01}&	\uline{87.79}\\
\textit{\textbf{Avg}} & 80.09 &	80.95 &	-	&\uline{83.85}& 	\textbf{84.88} 
 \\ \midrule
MSRA & 94.95 & \uline{95.97} & 87.99 & 92.22 & \textbf{96.01} \\
ResumeNER & \uline{95.93} & - & 93.92 & 95.90 & \textbf{96.05} \\ 
\textbf{\textit{Avg}} & \uline{94.72} & - & 90.96 & 94.06 & \textbf{96.03} \\
\bottomrule
\end{tabular}
}
\caption{Results of supervised evaluation on NER. Results on Chinese benchmarks are from~\citet{li-etal-2020-flat}.}
\label{tab:sup_ner}
\vspace{-4mm}
\end{table}

\subsection{Supervised Evaluation}\label{sec:supervised-evaluation}

The results for NER, RE, EE (including ED and EAE) tasks are shown in Tables~\ref{tab:sup_ner}, \ref{tab:sup_re}, and \ref{tab:sup_ee} respectively. 
\modelname outperforms the SoTA baselines on most benchmarks for four tasks and ranks within the top-$2$ results across all RE and ED benchmarks.
In the English IE, \modelname has achieved significant average improvements of $3.03$ and $2.92$ F1 points on the RE and ED tasks, with an improvement of $7.85$ points on the kbp37 of RE.
In the Chinese IE, \modelname has consistently achieved SoTA, with $4.12$ points average improvement over the SoTA baseline on the EAE task.
Moreover, it can be observed that the model not only achieves significant improvements in NER but also exhibits even greater enhancements in RE, ED, and EAE.
This indicates that the alignment phase enhances the multilingual NER capability through cross-lingual alignment, which improves the performance of all tasks by leveraging the foundational spotting capabilities of NER.

\begin{table}[!tbp]
\centering
\resizebox{.88\columnwidth}{!}{
\begin{tabular}{@{}lccc@{}}
\toprule
Dataset & YAYI-UIE & IEPILE & \modelname \\ \midrule
ADE corpus & \uline{84.14} & 83.73 & \textbf{84.45} \\
CoNLL 2004 & \textbf{79.73} & 72.87 & \uline{73.14} \\
GIDS & 72.36 & \uline{74.71} & \textbf{76.19} \\
kbp37 & 59.35 & \uline{65.09} & \textbf{72.94} \\
NYT & 89.97 & \uline{93.00} & \textbf{96.08} \\
NYT11-HRL & \textbf{57.53} & 53.19 & \uline{56.79} \\
SciERC & 40.94 & \uline{43.53} & \textbf{44.93} \\
Semeval RE & \uline{61.02} & 58.47 & \textbf{64.79} \\
\textbf{\textit{Avg}} &  \uline{68.13}  &  68.07  &  \textbf{71.16}  \\ \midrule
CMeIE & - & \uline{49.16} & \textbf{52.37} \\
DuIE 2.0 & \uline{81.19} & 75.61 & \textbf{82.85} \\
\textbf{\textit{Avg}} &  - &  62.39  &  \textbf{67.26}  \\ \bottomrule
\end{tabular} }
\caption{Results of supervised evaluation on RE.}
\label{tab:sup_re}
\end{table}

\begin{table}[!tbp]
\centering
\resizebox{.88\columnwidth}{!}{
\begin{tabular}{@{}llcccc@{}}
\toprule
Task & Dataset & BERT & YAYI-UIE & IEPILE & \modelname \\ \midrule
\multirow{7}{*}{ED} & ACE 2005 & \uline{72.50} & 65.00 & 72.46 & \textbf{73.57} \\
& CASIE & \textbf{68.98} & 63.00 & 60.07 & \uline{63.91} \\
& PHEE & - & 63.00 & \uline{63.22} & \textbf{67.03} \\
& \textbf{\textit{Avg}} & - & 63.67 & \uline{65.25} & \textbf{68.17} \\ \cmidrule(l){2-6} 
& DuEE 1.0 & 82.18 & 85.00 & \uline{86.73} & \textbf{87.18} \\
& DuEE-Fin & \uline{84.53} & 82.50 & 83.54 & \textbf{85.13} \\
& \textbf{\textit{Avg}} & 83.36 & 83.75 & \uline{85.14} & \textbf{86.16} \\
\bottomrule
\multirow{7}{*}{EAE} & ACE 2005 & 59.90 & 62.71 & \uline{63.90} & \textbf{69.95} \\
& CASIE & 60.37 & \uline{64.23} & 56.07 & \textbf{64.96} \\
& PHEE & - & \textbf{77.19} & 70.85 & \uline{76.24} \\
& \textbf{\textit{Avg}} & - & \uline{68.04} & 63.61 & \textbf{70.38} \\ \cmidrule(l){2-6} 
& DuEE 1.0 & 70.68 & \uline{78.08} & 75.63 & \textbf{82.12} \\
& DuEE-Fin & 75.73 & 70.02 & \uline{79.34} & \textbf{81.09} \\ 
& \textbf{\textit{Avg}} & 73.21 & 74.05 & \uline{77.49} & \textbf{81.61} \\
\bottomrule
\end{tabular}
}
\caption{Results of supervised evaluation on EE.
}
\label{tab:sup_ee}
\vspace{-4mm}
\end{table}

To further demonstrate that \modelname improves multilingual IE across all languages through IE cross-lingual alignment, we also compared it comprehensively with SoTA monolingual (English) IE systems, which are fine-tuned on LLMs including
InstructUIE~\cite{wang2023instructuie}, 
UniversalNER~\cite{zhouuniversalner}, GoLLIE~\cite{sainz2024gollie}, KnowCoder~\cite{li2024knowcoder}, GLiNER~\cite{zaratiana-etal-2024-gliner}, and GNER~\cite{ding2024rethinking}.
Appendix~\ref{appx:full-comparison} shows a detailed comparison with all baselines.
Compared to $10$ SoTA methods in the supervised setting, it achieves top-$2$ results on $14$ (of $23$) English NER benchmarks, indicating substantial enhancements.
Specifically, when compared to other code-based IE systems, such as KnowCoder and GoLLIE, \modelname still achieves a notable improvement.
This observation underscores the effectiveness of cross-lingual alignment of \modelname.
This indicates that our cross-lingual alignment not only avoids conflicts between different languages but also enhances overall multilingual IE performance by leveraging cross-lingual alignment.
Appendix~\ref{appx:computational-resource-comparison} shows a comparison of the computational resources.

\begin{table}[!t]
\centering
\resizebox{\columnwidth}{!}{
\begin{tabular}{@{}lcc@{}}
\toprule
 Dataset & Text-Based & Code-Based \\ \midrule
  ACE 2005~\cite{ace2005-annotation} & 84.93 & \textbf{86.07} \\
  AnatEM~\cite{DBLP:journals/bioinformatics/PyysaloA14} & 89.63 & \textbf{90.14} \\
   BC2GM~\cite{DBLP:conf/icpr/KocamanT20} & 82.53 & \textbf{83.17} \\
  BC5CDR~\cite{DBLP:journals/biodb/LiSJSWLDMWL16} & 90.03 & \textbf{90.17} \\
  CoNLL 2003~\cite{conll_2003} & 94.67 & \textbf{94.93} \\
  WNUT 2017~\cite{derczynski-etal-2017-results} & \textbf{66.14} & 66.03 \\ \midrule
  ResumeNER~\cite{zhang-yang-2018-chinese} & 94.31 & \textbf{95.37} \\
  MSRA~\cite{DBLP:conf/acl-sighan/Levow06} & 92.97 & \textbf{94.03} \\ \midrule
 \textbf{\textit{Avg}} & 86.90 & \textbf{87.49} \\ \bottomrule
\end{tabular}}
\caption{Ablation study on code-based multilingual IE.}
\label{tab:ablation_code}
\end{table}

\begin{table}[!t]
\centering
\resizebox{\columnwidth}{!}{
\begin{tabular}{@{}lccccc@{}}
\toprule
\textbf{Model} & \textbf{NER} & \textbf{RE} & \textbf{ED} & \textbf{EAE} & \textbf{\textit{Avg}} \\ \midrule
\modelname (w/o align) & 85.06 & 68.67 & 74.12 & 74.00 & 75.46\\
\modelname & \textbf{85.63 } & \textbf{70.38} & \textbf{75.37} & \textbf{74.87} & \textbf{76.56} \\
\bottomrule
\end{tabular}}
\caption{Ablation study on IE alignment training.}
\label{tab:ablation_align}
\vspace{-4mm}
\end{table}

\subsection{Ablation Study}

We conduct comprehensive ablation experiments to investigate whether the code-based multilingual IE and IE cross-lingual alignment phase of \modelname contribute to the performance.

\noindent \textbf{Code-Based Multilingual IE} \ \ 
Table~\ref{tab:ablation_code} shows results.
We evaluate the text-based~\cite{wang2023instructuie} and code-based multilingual IE method in the supervised setting.
We use $6$ English NER datasets, ACE 2005, AnatEM, BC2GM, BC5CDR, CoNLL 2003, WNUT 2017, and $2$ Chinese NER datasets, MSRA,   ResumeNER to conduct instruction tuning on the Baichuan2-7B for supervised evaluation.
We remove class comments in our schemas to ensure a fair comparison.
The code-based method outperforms by $0.59$ points, with improvements exceeding $1.00$ points in $3$ datasets, particularly in the Chinese datasets MSRA and ResumeNER.
The results show that the code can mitigate schema differences between languages, thereby enhancing the cross-lingual alignment to improve the performance of multilingual IE.

\noindent \textbf{IE Alignment Training} \ \ 
To show whether the IE cross-lingual alignment phase contributes to the performance, we further conduct an ablation study by removing the cross-lingual alignment phase of \modelname and denoting it as \modelname~\texttt{(w/o align phase)}.
Table~\ref{tab:ablation_align} shows the average results in the supervised setting on $4$ tasks across $45$ benchmarks.
Compared to \modelname~\texttt{(w/o align phase)}, the results indicate that \modelname achieves a substantial average improvement of $1.1$ points across four tasks.
This suggests that the IE cross-lingual alignment phase enhances the overall performance of multilingual IE, demonstrating its contribution to facilitating cross-lingual transfer within the multilingual IE framework.

\normalem

\begin{table}[tbp]
\centering
\resizebox{\columnwidth}{!}{
\begin{tabular}{@{}lccc@{}}
\toprule
Language & Text-Based & Code-Based & Code-Based~(w/ com.) \\ \midrule
German & 35.16 & 38.69 & \textbf{48.02} \\
Spanish & 43.19 & 48.93 & \textbf{52.16} \\
Dutch & 39.16 & 42.19 & \textbf{46.79} \\
Russian & 33.12 & 36.19 & \textbf{39.68} \\
Bengali & 18.23 & 17.59 & \textbf{23.95} \\
Persian & 23.45 & 24.69 & \textbf{28.54} \\
Hindi & 21.71 & 23.91 & \textbf{28.13} \\
Korean & 28.31 & 28.65 & \textbf{31.42} \\
Turkish & 32.36 & 35.16 & \textbf{41.16} \\ \midrule
\textit{\textbf{Avg}} & 30.52 & 32.89 & \textbf{37.76} \\ \bottomrule
\end{tabular}}
\caption{Analysis on cross-lingual transfer.}
\label{tab:analysis_code}
\vspace{-4mm}
\end{table}

\subsection{Analysis on Cross-Lingual Transfer}
To further investigate the impact of code-based unified schema representation in cross-lingual IE, we conduct cross-lingual evaluation finetuned on all NER datasets under three settings: text-based, code-based, and code-based~(with comments).
Table~\ref{tab:analysis_code} shows the results.
It indicates that adopting the code-based multilingual IE method can enhance cross-lingual performance by $2.37\%$.
This improvement can be attributed to the standardized representation of the same ontology across different languages provided by the code-based multilingual IE method.
Additionally, the integration of guideline information further improves cross-lingual performance by $4.85\%$, particularly in low-resource languages such as Bengali and Persian.
The code-based multilingual IE effectively facilitates the integration of guideline information, significantly enhancing cross-lingual IE performance.

\section{Conclusions}

We introduced \modelname, a code-based multilingual IE model that significantly enhances cross-lingual transfer through unified schema representation and IE cross-lingual alignment training.
\modelname utilizes Python classes to represent schemas uniformly across languages, ensuring semantic consistency in multilingual IE.
Additionally, the IE cross-lingual alignment phase leverages high-quality IE parallel datasets \datasetname, which are generated by our proposed LLM-based pipeline, to boost multilingual generalization.
Comprehensive experiments demonstrate that \modelname achieves SoTA in Chinese and English IE under various settings, with remarkable cross-lingual capability.

\section*{Limitation}

Expanding training to include languages beyond English and Chinese remains an area for further exploration.
Moreover, incorporating additional languages would provide valuable insights into the cross-lingual alignment mechanisms.
Additionally, addressing domain specificity will be a key focus of future research. Specialized vocabulary and domain-specific language use present significant challenges for cross-lingual transfer, making it essential to overcome these obstacles in upcoming work.
Introducing parallel data from other IE tasks during the cross-lingual alignment phase also offers promising avenues for further investigation.

\section*{Acknowledgments}

This work is partially funded by the National Natural Science Foundation of China under grants 62306299 and 62441229, the Lenovo-CAS Joint Lab Youth Scientist Project, the project under Grants No. JCKY2022130C039, and the Strategic Priority Research Program of the CAS under Grants No. XDB0680102. We thank anonymous reviewers for their insightful comments and suggestions.


\normalem
\bibliography{custom}

\clearpage

\appendix

\section{Description Generation}\label{appx:description-generation}

We provide a detailed explanation of the description generation process using the NER task as an example.
The generation process consists of two main phases, i.e., \textit{Description Initialization} and \textit{Description Polish}:
\begin{itemize}
    \item \textit{Description Initialization:} In this phase, we sample some entities to instruct the LLMs to summarize an initial description of these entities. The details are as follows:
    1) Randomly sample $10$ entities from the training set.
    2) Format the $10$ entities into the description initialization prompt template as shown in Tables~\ref{tab:description-init}, instructing the LLM to summarize and generate an initial description.
    \item \textit{Description Polish:} Based on the description generated by the first phase, this phase further instructs the LLM to polish the description based on a set of entities iteratively.
\end{itemize}

\textit{Description Initialization} often struggles to accurately represent and capture the characteristics of each entity. To address this, we introduce \textit{Description Polish}, a method designed to refine the details of the description, thereby enabling it to summarize the entities within a given entity type more effectively.

\section{Pipeline Evaluation}\label{app:pipeline-evaluation}

We evaluate our pipeline on the label projection benchmark WikiANN~\cite{wikiann}, which is a multilingual NER dataset.
We use $10$ languages, including Bengali~(bn), German~(de), Spanish~(es), Persian~(fa), Hindi~(hi), Korean~(ko), Dutch~(nl), Russian~(ru), Turkish~(tr), and Chinese~(zh), and randomly sample $1,000$ samples for each language.
We compare our pipeline with CLaP~\cite{parekh-etal-2024-contextual}, Awesome-Align~\cite{dou-neubig-2021-word}, and EasyProject~\cite{chen-etal-2023-frustratingly}.
We mainly evaluate faithfulness following the setting of CLaP, measured as the percentage of instances where translated labels appear in the translated sentence, as accuracy evaluation requires costly native speaker rankings across methods, incurring high costs.

Table~\ref{tab:label_projection} shows the result. We achieve $99\%$ in faithfulness on average, outperforming the current SoTA Awesome-Align by $3\%$ points on average.
We attribute the achievement to the latter two stages, which rephrase the span with minimal disturbance to the sentence.
Besides, other baselines experience significant performance drops in certain languages, such as EasyProject in Russia. In contrast, our pipeline achieves scores exceeding $93\%$ across all languages, which strongly demonstrates the robustness and reliability of our pipeline.
The results further substantiate the stability of our pipeline, demonstrating its capability to adapt effectively across different linguistic contexts.

\begin{table}[!tbp]
\centering
\resizebox{.95\columnwidth}{!}{
    \begin{tabular}{@{}c|ccc|c@{}}
\toprule
Language & Awesome-Align & EasyProject & CLaP & \modelname \\ \midrule
\textbf{bn} & 92 & 98 & 93 & \textbf{99} \\
\textbf{de} & 99 & 97 & 79 & \textbf{99} \\
\textbf{es} & 99 & 99 & 84 & \textbf{100} \\
\textbf{fa} & 96 & 99 & 72 & \textbf{100} \\
\textbf{hi} & 93 & 36 & 90 & \textbf{99} \\
\textbf{ko} & 96 & 93 & 64 & \textbf{99} \\
\textbf{nl} & 99 & 100 & 85 & \textbf{100} \\
\textbf{ru} & 97 & \textbf{99} & 66 & 93 \\
\textbf{tr} & 98 & 98 & 94 & \textbf{99} \\
\textbf{zh} & 92 & 92 & 60 & \textbf{99} \\ \midrule
\textbf{\textit{Avg}} & 96 & 91 & 79 & \textbf{99} \\ \bottomrule
\end{tabular}
}
\caption{Results of label projection evaluation.}
\label{tab:label_projection}
\end{table}

\section{Cross-Lingual Evaluation on Low Resource Languages}\label{appx:cross-ood-low}
Table~\ref{tab:cross-ood-low} shows results on the MasakhaNER 2.0.
\modelname achieves the best performance across a wider variety of low-resource languages.
Compared to IEPILE and GLiNER, \modelname achieves improvements of $11.41\%$ and $27.11\%$, respectively. These findings highlight the remarkable cross-lingual performance of \modelname on low-resource languages.
The results demonstrate that \modelname exhibits excellent cross-lingual generalization for IE, which can be attributed to the enhancement of IE cross-lingual alignment.

\begin{table}[ht]
\centering
\resizebox{\columnwidth}{!}{
\begin{tabular}{@{}ccccc@{}}
\toprule
\multicolumn{1}{c}{Language} & \multicolumn{1}{c}{YAYI-UIE} & \multicolumn{1}{c}{IEPILE} & \multicolumn{1}{c}{GLiNER} & \multicolumn{1}{c}{\modelname} \\ \midrule
Bamanankan & 31.19 & 29.00 & 28.85 & \textbf{32.97} \\
Ghomala & 32.03 & 32.36 & \textbf{34.71} & 32.97 \\
Éwé & 63.01 & 57.83 & 64.64 & \textbf{66.05} \\
Fon & 30.05 & 29.73 & 30.58 & \textbf{31.86} \\
Hausa & 47.68 & 43.06 & 42.04 & \textbf{48.35} \\
Igbo & 37.14 & 37.54 & 36.08 & \textbf{38.75} \\
Kinyarwanda & 40.84 & 43.87 & 45.04 & \textbf{47.56} \\
Luganda & 46.74 & 53.13 & 48.59 & \textbf{58.04} \\
Luo & 44.04 & 48.02 & 39.04 & \textbf{49.89} \\
Mossi & 26.05 & 26.42 & 30.96 & \textbf{31.71} \\
Chichewa & \textbf{53.72} & 52.93 & 48.52 & 51.94 \\
Nigerian-Pidgin & 66.71 & 67.50 & 39.96 & \textbf{81.91} \\
Shona & 43.47 & 43.87 & 34.37 & \textbf{46.58} \\
Swahili & 64.62 & 55.50 & 58.06 & \textbf{69.42} \\
Setswana & 51.45 & \textbf{54.05} & 39.83 & \textbf{56.97} \\
Twi & 35.62 & 45.69 & 28.54 & \textbf{51.10} \\
Wolof & 34.75 & 35.31 & 36.27 & \textbf{38.70} \\
Xhosa & 36.57 & 36.99 & 29.25 & \textbf{45.73} \\
Yoruba & 28.88 & 26.79 & 5.06 & \textbf{34.73} \\
Zulu & \textbf{38.07} & 34.61 & 28.34 & 36.50 \\ \midrule
\textbf{\textit{Avg}} & 42.63 & 42.71 & 37.44 & \textbf{47.59$^{\uparrow\textbf{11.43\%}}$} \\ \bottomrule
\end{tabular}
}
\caption{Results of cross-lingual evaluation on MasakhaNER 2.0.}
\label{tab:cross-ood-low}
\vspace{-4mm}
\end{table}

\section{Zero-Shot Evaluation}\label{appx:zero-shot}

\paragraph{Datasets} For the zero-shot evaluation, among NER benchmarks, we use $5$ English benchmarks from CrossNER~\cite{DBLP:conf/aaai/Liu0YDJCMF21} and $2$ Chinese benchmarks, Weibo~\cite{DBLP:conf/emnlp/PengD15} and Boson\footnote{\url{https://github.com/InsaneLife/}}.
Among RE benchmarks, we adopt $2$ English benchmarks FewRel~\citep{DBLP:conf/emnlp/HanZYWYLS18}, Wiki-ZSL~\citep{DBLP:conf/naacl/ChenL21} and $3$ Chinese benchmarks, COAE2016\footnote{\url{https://github.com/Sewens/COAE2016}\label{fn:coae2016}}, IPRE~\citep{wang2019ipre}, and SKE2020\footnote{\url{https://aistudio.baidu.com/datasetdetail/177191}\label{fn:SKE2020}}.
Among ED benchmarks, we use $3$ English benchmarks CrudeOil News~\cite{lee2021annotatedcommoditynewscorpus}, RAMS~\cite{ebner-etal-2020-multi}, WikiEvents~\cite{DBLP:conf/naacl/LiJH21} and $2$ Chinese benchmarks FewFC~\citep{DBLP:conf/aaai/Zhou0ZWXL21} and CCF law\footnote{\url{https://aistudio.baidu.com/projectdetail/4201483}}.
\paragraph{Results} Tables~\ref{tab:zero_en} and \ref{tab:zero_zh} show the zero-shot performance in English and Chinese, respectively.
\modelname achieved SoTA performance across all benchmarks in Chinese tasks, demonstrating substantial and consistent improvements in generalization ability.
In English tasks, \modelname exhibited a noteworthy average increase of $4.24$ points on the NER task.
Similarly, in Chinese tasks, \modelname surpassed existing SoTA models on all benchmarks, achieving an average improvement of $3.29$ points.
Moreover, the trends demonstrated in both Chinese and English show that the most significant improvement occurs in NER.
We hypothesize that the NER alignment training in the first phase is more challenging to generalize to other IE tasks in out-of-domain scenarios, which leaves the introduction of RE and EE alignment training to future work.

\begin{table}[tbp]
\centering
\resizebox{0.87\columnwidth}{!}{
\begin{tabular}{@{}cccccc@{}}
\toprule
Task & Dataset & GPT-4 & YAYI-UIE & IEPILE & \modelname\\ \midrule
\multirow{6}{*}{NER} & AI & - & 52.40 & - & \textbf{65.04} \\
 & Literature & - & 45.99 & - & \textbf{59.68} \\
 & Music & - & 51.20 & - & \textbf{62.97} \\
 & Politics & - & 51.82 & - & \textbf{63.65} \\
 & Science & - & 50.53 & - & \textbf{62.29} \\
 & \textit{\textbf{Avg}} & 58.49 & 50.39 & 55.55 & \textbf{62.73} \\ \midrule
\multirow{2}{*}{RE} & FewRel & 22.43 & 36.09 & 41.28 & \textbf{42.18} \\
 & Wiki-ZSL & 23.76 & 41.07 & 37.61 & \textbf{42.13} \\ \midrule
\multirow{3}{*}{ED} & CrudeOil News & 26.13 & 12.45 & \textbf{ 36.61}& 33.47 \\
& WikiEvents & 5.24 & 10.97 & 9.12 & \textbf{12.59} \\
& RAMS& 10.14 & 18.87 & 20.19& \textbf{20.71} \\\midrule
\textit{\textbf{Avg}} & - & 38.02 			 
 & 37.14 & 42.26& \textbf{46.47} \\ \bottomrule
\end{tabular} }
\caption{Results of zero-shot evaluation on English Benchmarks. Results of GPT-4 are from~\citet{gui-etal-2024-iepile}}
\label{tab:zero_en}
\end{table}

\begin{table}[tbp]
\centering
\resizebox{0.89\columnwidth}{!}{
\begin{tabular}{@{}cccccc@{}}
\toprule
Task & Dataset & GPT-4 & YAYI-UIE & IEPILE & \modelname \\ \midrule
\multirow{2}{*}{NER} & Boson & 48.15 & 49.25 & \uline{55.77} & \textbf{59.58} \\
 & Weibo & 29.80 & \uline{38.03} & 36.46 & \textbf{44.01}$^\dag$ \\ \midrule
\multirow{3}{*}{RE} & COAE2016 & 41.15 & 19.97 & \uline{47.43} & \textbf{48.79} \\
 & IPRE & 18.15 & 22.97 & \uline{29.76} & \textbf{32.43} \\
 & SKE2020 & 56.77 & 70.80 & \uline{72.50} & \textbf{72.91} \\ \midrule
\multirow{2}{*}{ED} & CCF Law & 42.12 & 12.87 & \uline{63.53} & \textbf{68.90} \\
 & FewEC & 74.25 & 81.28 & \uline{83.59} & \textbf{84.87} \\ \midrule
\textit{\textbf{Avg}} & - & 44.34 & 42.17 & \uline{55.58} & \textbf{58.78} \\ \bottomrule
\end{tabular} }
\caption{Results of zero-shot evaluation on Chinese Benchmarks.
$^\dag$ donates the result of Weibo using four types following the setups of the two baseline models, and the result for the eight types is $38.80$.}
\label{tab:zero_zh}
\end{table}

\begin{table*}[t!]
\centering
\resizebox{1.0\textwidth}{!}{
\begin{tabular}{@{}c|cccccccccc|c@{}}
\toprule
Dataset & BERT & InsturctUIE & IEPILE & YAYI-UIE & GoLLIE & UniversalNER & GLiNER & KnowCoder & GNER & B\textsuperscript{2}NER & \modelname \\ \midrule
ACE 2005 & 87.30 & 79.94 & 81.86 & 81.78 & \textbf{89.10} & 86.69 & 82.80 & 86.10 & - & 83.04 & \uline{87.49} \\
AnatEM & 85.82 & 85.82 & 87.21 & 76.54 & - & 88.65 & 88.90 & 86.40 & \textbf{90.24} & 89.18 & \uline{89.19} \\
BC2GM & 80.90 & 80.69 & 80.73 & 82.05 & - & 82.42 & \uline{83.70} & 82.00 & 83.18 & 81.95 & \textbf{84.49} \\
BC4CHEMD & 86.72 & 87.62 & \textbf{90.45} & 88.46 & - & 89.21 & 87.90 & - & 89.40 & 88.96 & \uline{89.57} \\
BC5CDR & 85.28 & 89.02 & 88.07 & 83.67 & \textbf{91.90} & 89.34 & 88.70 & 89.30 & \uline{90.27} & 88.52 & 88.46 \\
Broad Twitter & 58.61 & 80.27 & \uline{83.52} & - & - & 81.25 & 82.50 & 78.30 & \textbf{83.74} & 82.16 & 82.36 \\
CoNLL 2003 & 92.40 & 91.53 & 92.49 & \textbf{96.77} & 92.90 & 93.30 & 92.60 & 94.10 & 93.60 & 92.56 & \uline{94.69} \\
FabNER & 64.20 & 78.38 & 77.07 & 72.63 & - & 81.87 & 77.80 & 82.90 & \textbf{85.39} & 78.82 & \uline{83.19} \\
FindVehicle & 87.13 & 87.56 & 98.49 & 98.47 & - & 98.30 & 95.70 & \uline{99.40} & 98.62 & 97.89 & \textbf{99.47} \\
GENIA & 73.30 & 75.71 & 76.66 & 75.21 & - & 77.54 & \uline{78.90} & 76.70 & - & 76.43 & \textbf{78.97} \\
HarveyNER & \textbf{80.20} & 74.49 & 67.70 & 69.57 & - & 74.21 & - & - & \uline{74.73} & 73.67 & 73.91 \\
MIT Movie & 88.78 & 89.58 & 88.23 & 70.14 & - & 90.17 & 87.90 & 89.70 & 90.23 & \textbf{90.78} & \uline{90.24} \\
MIT Restaurant & 81.02 & 82.59 & 79.85 & 79.38 & - & 82.35 & \uline{83.60} & 81.30 & 81.73 & \textbf{83.71} & 81.95 \\
MultiNERD & 91.25 & 90.26 & 94.60 & 88.42 & - & 93.73 & 93.80 & \textbf{96.10} & 94.30 & 93.98 & \uline{95.94} \\
NCBI & 80.20 & 86.21 & 85.26 & 87.29 & - & 86.96 & \uline{87.80} & 83.80 & \textbf{89.27} & 84.83 & 85.49 \\
OntoNotes & \textbf{91.11} & 88.64 & 87.04 & 87.55 & - & 89.91 & - & - & \uline{90.69} & 84.31 & 87.91 \\
Polyglot & \textbf{75.65} & 53.30 & \uline{70.85} & - & - & 65.70 & 61.50 & - & 67.52 & 61.96 & 64.47 \\
TweetNER7 & 56.49 & 65.90 & \uline{66.99} & - & - & 65.80 & 51.40 & - & 66.87 & 66.26 & \textbf{67.98} \\
WikiANN & 70.60 & 64.47 & 72.63 & - & - & 84.91 & 83.70 & \textbf{87.70} & \uline{86.87} & 85.07 & 84.69 \\
WikiNeural & 82.78 & 88.27 & 87.63 & - & - & \uline{93.28} & 91.30 & - & \textbf{93.71} & 93.01 & 87.79 \\
OntoNotes 5 & 84.28 & - & - & - & 83.40 & - & - & \uline{88.20} & - & - & \textbf{88.35} \\
DIANN & 82.05 
 & - & - & - & 84.10 & - & - & \uline{94.70} & - & - & \textbf{94.71} \\
WNUT 2017 & 36.16 & - & - & - & 53.70 & - & - & \uline{66.40} & - & - & \textbf{68.72} \\ \bottomrule
\end{tabular} }
\caption{Full comparison of supervised evaluation on NER. The results of BERT on DIANN, OntoNotes 5, and WNUT 2017 are from our implementation.}
\label{tab:full-ner}

\end{table*}

\section{Full Comparison of Supervised Evaluation}\label{appx:full-comparison}

We conduct a comprehensive supervised comparison in English between \modelname and all current SoTA methods, including monolingual IE methods and multilingual IE methods under the supervised setting in Tables~\ref{tab:full-ner},~\ref{tab:full-re}, and~\ref{tab:full-ee}.
In the NER task, we demonstrate top-$2$ results across $14$ (of $23$)  benchmarks; in the RE task, we achieve  top-$2$ results across $5$ (of $8$) benchmarks.
Moreover, particularly in the ED and EAE tasks, we rank among the top-$2$ results across all benchmarks, further substantiating the significant efficacy of our method in multilingual IE.

Additionally, in the comparison of three benchmarks DIANN~\cite{Zavala2018AHB}, Ontonotes 5~\cite{OntoNotes_Dataset}, and WNUT 2017~\cite{derczynski-etal-2017-results}, utilized by other English code-based IE work~\cite{sainz2024gollie,li2024knowcoder}, \modelname outperforms the other baselines, which further demonstrated that cross-lingual alignment can enhance monolingual IE.

\begin{table}[!tbp]
\centering
\resizebox{\linewidth}{!}{
\begin{tabular}{@{}c|cccc|c@{}}
\toprule
Dataset  & YAYI-UIE & IEPILE & InsturctUIE & KnowCoder & \modelname \\ \midrule
ADE corpus  & 84.14 & 83.73 & 82.31 & \uline{84.30} & \textbf{84.45} \\
CoNLL 2004  & \textbf{79.73} & 72.87 & \uline{78.48} & 73.30 & 73.14 \\
GIDS & 72.36 & 74.71 & \uline{76.90} & \textbf{78.00} & 76.19 \\
kbp37  & 59.35 & 65.09 & 36.14 & \textbf{73.20} & \uline{72.94} \\
NYT & 89.97 & 93.00 & 91.00  & \uline{93.70} & \textbf{96.08} \\
NYT11-HRL  & \textbf{57.53} & 53.19  & 56.06 & - & \uline{56.79} \\
SciERC  & 40.94 & \uline{43.53}  & 37.40 & 40.00 & \textbf{44.93} \\
Semeval RE  & 61.02 & 58.47  & \textbf{73.23} & \uline{66.30} & 64.79 \\ \bottomrule
\end{tabular}
}
\caption{Full comparison of supervised evaluation on RE.}
\label{tab:full-re}
\end{table}

\section{Computational Resource Comparison}\label{appx:computational-resource-comparison}
In the Universal IE Training Phase, we only train on datasets of the supervised setting, which is the same as the other baselines and necessary for fair comparison.
Therefore, the additional computational resources of \modelname are mainly from the IE Cross-Lingual Alignment Phase.
First, this is one of our core contributions, which necessitates additional computational resources. Second, this phase of training is highly efficient.
In this phase, we train the model for one epoch with $257$k samples, which is considered a reasonable computational cost.
For comparison, the additional computational resources required for the Schema Understanding Training of KnowCoder is one epoch with $926$k samples ($\sim$4x).
Meanwhile, KnowCoder even has a large-scale pre-training phase.

\section{Data Statistics}\label{appx:data-statics}

\subsection{\datasetname}\label{app:parallel}
In this section, we mainly introduce the statistics of \datasetname, which is constructed from two datasets WikiNeural~\cite{tedeschi2021wikineural} and CLUENER2020~\cite{xu2020cluener2020}. The detailed statistics are shown in Table~\ref{tab:dataset-parallel}.

\subsection{Dataset Statistics}
In this work, we conduct evaluations on $64$ datasets, comprising $34$ datasets for the NER task, $15$ datasets for the RE task, $10$ datasets for the ED task, and $5$ datasets for the EAE task.
To fairly compare with code-based methods, KnowCoder and GoLLIE in demonstrating the effectiveness of our cross-lingual alignment, we also used PileNER~\cite{zhouuniversalner} for training.
The detailed statistics are shown in Tables~\ref{tab:dataset-ner}, \ref{tab:dataset-re}, and \ref{tab:dataset-ee}, respectively.

\begin{table}[!t]
\centering
\resizebox{\linewidth}{!}{
\begin{tabular}{@{}cc|cccc|c@{}}
\toprule
Task & Dataset & BERT & YAYI-UIE & IEPILE & KnowCoder & \modelname \\ \midrule
\multirow{3}{*}{ED} & ACE 2005 & 72.50 & 65.00 & 72.46 & \textbf{74.2} & \uline{73.57} \\
 & CASIE & \textbf{68.98} & 63.00 & 60.07 & - & \uline{63.91} \\
 & PHEE & - & 63.00 & \uline{63.22} & - & \textbf{67.03}\\ \midrule
\multirow{3}{*}{EAE} & ACE 2005 & 59.90 & 62.71 & 63.90 & \textbf{70.30} & \uline{69.95} \\
 & CASIE & 60.37 & \uline{64.23} & 56.07 & - & \textbf{64.96} \\
 & PHEE & - & \textbf{77.19} & 70.85 & - & \uline{76.24} \\ \bottomrule
\end{tabular}
}
\caption{Full comparison of supervised evaluation on ED and EAE.}
\label{tab:full-ee}
\end{table}

\section{Examples}\label{appx:examples}
We present the examples of instruction-tuning data for the IE cross-lingual alignment phase and the multi-lingual IE training phase in Figure \ref{fig:align_prompt} and Figure \ref{fig:multi_prompt}, respectively.

\section{Prompts}\label{appx:prompts}
\paragraph{IE Parallel Data Construction Pipeline}
Using en$\Rightarrow$zh as an example, we introduce the prompts we used in the IE parallel data construction pipeline including three stages \textit{Joint Translation}, \textit{Span Rephrase}, and \textit{Sentence Rephrase} in Tables~\ref{tab:translate-prompt},~\ref{tab:span-rephrase-prompt}, and~\ref{tab:sentence-rephrase-prompt}.

\begin{table*}[!h]
\centering
\resizebox{.97\textwidth}{!}{
\begin{tabular}{@{}cccccccc@{}}
\toprule
\textbf{\#Source}    & \textbf{\#Origin Language} & \textbf{\#Types} & \textbf{\#Instances} & \textbf{\#Train Num} & \textbf{\#Valid Num} & \textbf{\#Test Num} & \textbf{\#Total Num} \\ \midrule
WikiNeural  & En              & 3       & 149,011      & 92,720  & 11,590   & 11,597  & 115,907  \\
CLUENER2020 & Zh              & 10      & 19,720       & 10,000  & 1,343    & 1,345   & 12,688   \\ \bottomrule
\end{tabular}
}
\captionof{table}{Statistics of \datasetname datasets.}
\label{tab:dataset-parallel}
\end{table*}

\begin{table*}[!h]
\resizebox{\textwidth}{!}{
\begin{tabular}{cc|c|cc|ccc}
\toprule
\textbf{\#Dataset} & \textbf{\#Types} & \textbf{\#Major Domain} & \textbf{\#Train} & \textbf{\#Test} & \textbf{\#Train Num} & \textbf{\#Valid Num} & \textbf{\#Test Num}\\
\midrule

ACE2005~\cite{ace2005-annotation} & 7 & News & \checkmark & \checkmark & 7,299 & 964 & 1,060 \\
AnatEM~\cite{DBLP:journals/bioinformatics/PyysaloA14} & 1 & Biomedical & \checkmark & \checkmark & 5,861 & 2,081 & 3,830 \\
BC2GM~\cite{DBLP:conf/icpr/KocamanT20} & 1 & Biomedical & \checkmark & \checkmark & 12,500 & 2,500 & 5,000 \\

BC4CHEMD~\cite{DBLP:conf/icpr/KocamanT20} & 1 & Biomedical & \checkmark & \checkmark & 30,488  & 30,468  & 26,204 \\
BC5CDR~\cite{DBLP:journals/biodb/LiSJSWLDMWL16} & 2 & Biomedical & \checkmark & \checkmark & 4,560 & 4,581 & 4,797 \\

Broad Twitter~\cite{broad_twitter_corpus_DATASET} & 3 & Social Media & \checkmark & \checkmark & 5334   & 2,000 & 2,001\\

CoNLL2003~\cite{conll_2003} & 3 & News & \checkmark & \checkmark & 14,041 & 3,250 & 3,453 \\

FabNER~\cite{Kumar2021FabNERIE} & 12 & Science & \checkmark & \checkmark & 9,435 & 2,182 & 2,064 \\
FindVehicle~\cite{guan2024findvehicle} & 21 & Traffic & \checkmark & \checkmark & 21,547 & 20,777 & 20,769 \\

GENIA~\cite{GENIANER_DATASET} & 5 & Biomedical & \checkmark & \checkmark & 15,023 & 1,669 & 1,854 \\

HarveyNER~\cite{DBLP:conf/naacl/ChenXZH22} & 4 & Social Media & \checkmark & \checkmark & 3,553  & 1,270 & 1,260  \\

MIT Movie~\cite{MITReviewDataset} & 12 & Social Media & \checkmark & \checkmark & 9,774 & 2,442 & 2,442 \\
MIT Restaurant~\cite{MITReviewDataset} & 8 & Social Media & \checkmark & \checkmark & 7,659 & 1,520 & 1,520 \\
MultiNERD~\cite{multiNERD_DATASET} & 16 & Wikipedia & \checkmark & \checkmark & 134,144 & 10,000 & 10,000 \\
NCBI~\cite{DBLP:journals/jbi/DoganLL14} & 1 & Biomedical & \checkmark & \checkmark & 5,432  & 923  & 940 \\
OntoNotes~\cite{DBLP:conf/naacl/PradhanX09} & 18 & General & \checkmark & \checkmark & 107,032 & 14,110 & 10,838 \\

PolygtNER~\cite{polyglot-NER_DATASET} & 3 & Wikipedia & \checkmark &  \checkmark 
 & 393,941 & - & 10,000\\
TweetNER7~\cite{tweetNER7_DATASET} &  7   & Social Media & \checkmark &  \checkmark 
 & 1,325   & 7,111 & 576\\
 WikiANN~\cite{wikiann} &  3  & Wikipedia & \checkmark & \checkmark &  20,000  & - & 10,000 \\
 WikiNeural~\cite{tedeschi2021wikineural} &  3  & Wikipedia & \checkmark & \checkmark &  92,720 &11,590& 11,597  \\
DIANN~\cite{Zavala2018AHB} & 1 & Science & \checkmark & \checkmark & 3,900 & 975 & 1,334 \\
OntoNotes5~\cite{OntoNotes_Dataset} & 18 & General & \checkmark & \checkmark &  54994 & 7997 & 7782  \\
 WNUT 2017~\cite{derczynski-etal-2017-results} & 6 & General & \checkmark & \checkmark & 3,394 & 1,008 & 1,287 \\
PileNER~\cite{zhouuniversalner} & 15,578 & General & \checkmark &  & 45,883 & - & - \\

CrossNER\_AI~\cite{DBLP:conf/aaai/Liu0YDJCMF21} & 13 & AI &  & \checkmark & 100 & 350 & 431 \\
CrossNER\_literature~\cite{DBLP:conf/aaai/Liu0YDJCMF21} & 11 & Literary &  & \checkmark & 100 & 400 & 416 \\
CrossNER\_music~\cite{DBLP:conf/aaai/Liu0YDJCMF21} & 12 & Musical &  & \checkmark & 100 & 380 & 465 \\
CrossNER\_politics~\cite{DBLP:conf/aaai/Liu0YDJCMF21} & 8 & Political &  & \checkmark & 199 & 540 & 650 \\
CrossNER\_science~\cite{DBLP:conf/aaai/Liu0YDJCMF21} & 16 & Scientific &  & \checkmark & 200 & 450 & 543 \\ \midrule
MSRA~\cite{DBLP:conf/acl-sighan/Levow06} & 3 & News & \checkmark & \checkmark & 46,364 & 4,500 & 3,442 \\
ResumeNER~\cite{zhang-yang-2018-chinese} & 8 & News & \checkmark & \checkmark & 3,821 & - & 477 \\
Weibo~\cite{DBLP:conf/emnlp/PengD15} & 8 & Social media &  & \checkmark & 1,350 & - & 270 \\
Boson\footnote{\url{https://aistudio.baidu.com/datasetdetail/177191}} & 6 & News &  & \checkmark & - & - & 191 \\  \midrule
Multiconer\cite{malmasi-etal-2022-multiconer} & 6 & News &  & \checkmark & - & - & 51,793 \\  
MasakhaNER 2.0\cite{adelani2022masakhaner} & 4 & News &  & \checkmark & - & - & 30,538 \\  
\bottomrule

\end{tabular}}
\captionof{table}{Statistics of NER datasets.}
\label{tab:dataset-ner}
\end{table*}

\clearpage

\begin{table*}[!h]
        \resizebox{\textwidth}{!}{
        \begin{tabular}{cc|c|cc|ccc}
             \toprule
            \textbf{\#Dataset} & \textbf{\#Types} & \textbf{\#Major Domain} & \textbf{\#Train} & \textbf{\#Test} & \textbf{\#Train Num} & \textbf{\#Valid Num} & \textbf{\#Test Num}\\
    
            \midrule

ADE corpus~\cite{ADEcorpus_DATASET} & 5 & Biomedical & \checkmark & \checkmark & 3417 & 427 & 428 \\ 
CoNLL 2004~\cite{roth-yih-2004-linear} & 5 & News & \checkmark & \checkmark & 922 & 231 & 288 \\
GIDS~\cite{GIDS_DATASET} & 4 & News & \checkmark & \checkmark & 8526 & 1417 & 4307 \\
NYT~\cite{NYT_DATASET} & 24 & News & \checkmark & \checkmark & 56,196 & 5000 & 5000 \\
NYT11~\cite{NYT11_DATASET} & 12 & News & \checkmark & \checkmark & 60765 & - & 362 \\
kbp37~\cite{kbp37_DATASET} & 18 & News & \checkmark & \checkmark & 15917 & 1724 & 3405 \\
SciERC~\cite{SciERC_DATASET} & 7 & Scientific & \checkmark & \checkmark & 1861 & 275 & 551 \\
semeval RE~\cite{hendrickx-etal-2010-semeval} & 10 & Scientific & \checkmark & \checkmark & 6507 & 1493 & 2717 \\
FewRel~\cite{DBLP:conf/emnlp/HanZYWYLS18} & 100 & Wikipedia &  & \checkmark & - & - & 17291 \\
Wiki-ZSL~\cite{DBLP:conf/naacl/ChenL21} & 83 & Wikipedia &  & \checkmark & - & - & 23113 \\ \midrule
DuIE2.0~\cite{DBLP:conf/emnlp/LuanHOH18} & 48 & News & \checkmark & \checkmark & 171126 & 20652 & - \\
CMeIE~\cite{DBLP:conf/nlpcc/GuanZZXZ20} &  53   & Biomedical & \checkmark & \checkmark & 14339 & 3585 & - \\
COAE2016\footnote{\url{https://github.com/Sewens/COAE2016}} & 9 & General &  & \checkmark & - & - & 971 \\
IPRE~\cite{wang2019ipre} & 35 & General &  & \checkmark & - & - & 3340 \\
SKE2020\footnote{\url{https://aistudio.baidu.com/datasetdetail/177191}} & 49 & News &  & \checkmark & - & - & 3601 \\ \bottomrule
    
\end{tabular}}
\captionof{table}{Statistics of RE datasets. Since the test set for DuIE2.0 is not open-sourced, we use the valid set as our evaluation data.}
\label{tab:dataset-re}
\end{table*}

\begin{table*}[h]
\resizebox{\textwidth}{!}{
\begin{tabular}{cc|c|cc|ccc}
     \toprule
    \textbf{\#Dataset} & \textbf{\#Types} & \textbf{\#Major Domain} & \textbf{\#Train} & \textbf{\#Test} & \textbf{\#Train Num} & \textbf{\#Valid Num} & \textbf{\#Test Num}\\

    \midrule

ACE05~\cite{ace2005-annotation} & 33 & News & \checkmark & \checkmark & 19216 & 901 & 676 \\
CASIE~\cite{DBLP:conf/aaai/SatyapanichFF20} & 5 & Cybersecurity & \checkmark & \checkmark & 3732 & 777 & 1492 \\
PHEE~\cite{DBLP:conf/emnlp/Sun0PWJGK022} & 2 & Biomedical & \checkmark & \checkmark & 2897 & 960 & 968 \\
CrudeOilNews~\cite{lee2021annotatedcommoditynewscorpus} & 18 & Oil News &  & \checkmark & - & - & 356 \\
RAMS~\cite{ebner-etal-2020-multi} & 106 & News &  & \checkmark & - & - & 887 \\
WikiEvents~\cite{DBLP:conf/naacl/LiJH21} & 31 &Wikipedia &  & \checkmark & - & - & 249 \\\midrule
DuEE1.0~\cite{DBLP:conf/nlpcc/LiLPCPWLZ20} & 65 & News & \checkmark & \checkmark & 11908 & 1492 & - \\
DuEE-fin~\cite{DBLP:conf/nlpcc/HanZLXPZ22} & 13 & Finance & \checkmark & \checkmark & 7015 & 1171 & - \\
CCF\_law\footnote{\url{https://aistudio.baidu.com/projectdetail/4201483}} & 9 & Law &  & \checkmark & - & - & 971 \\
FewFC~\cite{DBLP:conf/aaai/Zhou0ZWXL21} & 5 & Finance &  & \checkmark & - & - & 2879 \\ \bottomrule

\end{tabular}}
\captionof{table}{Statistics of EE datasets. Since the test sets for DuEE1.0 and DuEE-fin are not open-sourced, we use the valid sets as our evaluation data. }
\label{tab:dataset-ee}
\end{table*}

\end{CJK}

\begin{figure*}[h]
   \centering
   \includegraphics[width=.98\linewidth]{align_instruction_tuning.pdf}
   \caption{An example of instruction-tuning data for the IE cross-lingual alignment phase.}
   \label{fig:align_prompt}
\end{figure*}

\begin{figure*}[h]
   \centering
   \includegraphics[width=0.98\linewidth]{multi_instruction_tuning.pdf}
   \caption{An example of instruction-tuning data for the multilingual IE training phase.}
   \label{fig:multi_prompt}
\end{figure*}

\begin{CJK}{UTF8}{gkai}
\begingroup
\begin{table*}[htp]
    \centering
    
    \begin{tabular}{p{\textwidth}}
        \toprule
        \underline{\textbf{\textsc{Prompt for Description Initialization.}}} \\
        \vspace{-2mm}
        \# Writing Entity Descriptions\\ \\ \#\# Introduction \\
        This guide provides a step-by-step process for writing a clear, concise, and accurate description of an entity type based on a provided list of examples. The objective is to generalize the shared characteristics of the examples without referencing any specific instance, giving a broad and comprehensive understanding of the entity type.\\ \\
        \#\# Prerequisites \\
        Before you begin, make sure you have the following: \\
        - **Entity Type**: The name of the entity type that requires a description. \\
        - **Entity List**: A set of examples representing this entity type. \\
        - **Basic Description**: While not mandatory, familiarity with the general concept of the entity type could be beneficial. \\ \\
        \#\# Step-by-Step Instructions  \\
        \#\#\# Step 1: Begin with the Required Phrase \\
        Each description should start with: \\
        **"[Entity Type] refers to"**  \\ 
        This ensures consistency across all descriptions. Replace **[Entity Type]** with the actual type name. \\ \\
        \#\#\# Step 2: Generalize the Shared Characteristics \\
        - Review the **Entity List** to identify common traits among all examples. \\
        - Avoid referring to specific examples directly. Generalize to cover the entire group. \\
        \vspace{-4mm}- Example: If the list includes various vehicles (cars, trucks), the description should focus on common traits such as modes of transportation designed for movement.

         \\

        \#\#\# Step 3: Provide Comprehensive Coverage \\
        - The description should encapsulate all critical aspects represented in the example list, accounting for any outliers or unusual cases. \\
            - Example: If the list includes motorized vehicles and non-motorized bicycles, ensure the description covers both. \\ \\
        \#\#\# Step 4: Output the Description \\
        - After completing the description, present it without referencing explicit examples. It should summarize the entity type in a single, generalized statement. \\
        - If you want to revise the description, output the modified description in the format. \\
        \\

        \#\# Conclusion \\
        By following these steps, you will create an accurate, clear, and generalized description of an entity type. Start with the required phrase, focus on generalization, and keep the language simple yet precise. \\ \\

        \#\# Input \\
        Entity Type: \{entity\_type\} \\
        Entity Example List: \{entity\_example\_list\} \\ \\
        \#\# Example Template for Output \\
        Entity Type: \{entity\_type\} \\
        Entity Example List: \{entity\_example\_list\} \\
        Entity Type Description: "\{entity\_type\} refers to..."(in the language of the Entity list) \\


        \bottomrule
    \end{tabular}
    
    \caption{Prompt for Description Initialization.}
    \label{tab:description-init}
\end{table*}

\endgroup
\end{CJK}

\begin{CJK}{UTF8}{gkai}
\begingroup
\begin{table*}[htp]
    \centering
    
    \begin{tabular}{p{\textwidth}}
        \toprule
        \underline{\textbf{\textsc{Prompt for Description Polish.}}} \\
        \vspace{-2mm}
        \# Evaluating and Revising Entity Description\\ \\ \#\# Introduction \\
        This guide provides a systematic approach to evaluate whether a given description accurately represents the characteristics of an entity type. If the description is accurate and complete, no revision is necessary. However, if inaccuracies or omissions exist, revisions are required to ensure clarity and consistency in classifying entities.\\ \\
        \#\# Step-by-Step Instructions \\
        \#\#\# Step 1: Analyze the Entity Type Description \\
        - Carefully review the **entity type description** provided.\\
        - Example: For the entity type ``Animal,'' the description may include ``living organisms that move, breathe, and consume organic matter.'' \\  \\
        \#\#\# Step 2: Analyze the Entity \\
        - Review the specific entity's characteristics, noting its unique features. \\
        - Example: If the entity is ``dog,'' note traits like ``mammal, four-legged, domesticated, etc.'' \\
         \\

        \#\#\# Step 3: Evaluate the Description's Accuracy and Completeness \\
        - Compare the entity type description with the entity's characteristics. \\
            \hspace{1cm}- Does the description fully encompass the defining features of the entity? \\
            \hspace{1cm}- Are any characteristics missing or misrepresented? \\ 
        - Check for completeness: \\    
            \hspace{1cm}- Does the description cover all essential traits necessary for classification? \\
        - Verify accuracy: \\    
            \hspace{1cm}- Are the described attributes factually correct? \\
            \\
        \#\#\# Step 4: Revise the Description (if necessary) \\
        - If the description is incomplete or inaccurate, revise it to reflect the entity's correct characteristics. \\ 
        - Ensure the revised description is clear, precise, and free from ambiguities. \\
        \\

        \#\# Conclusion \\
        Following these steps will ensure each entity's description is both accurate and comprehensive. This process maintains clarity and consistency in classifying entities under their respective types.
 \\ \\

        \#\# Input \\
        Entity Type: \{entity\_type\} \\
        Entity Example List: \{entity\_example\_list\} \\ \\
        \#\# Example Template for Output \\
        Entity Type: \{entity\_type\} \\
        Entity Example List: \{entity\_example\_list\} \\
        Entity Type Description: "\{entity\_type\} refers to..."(in the language of the Entity list) \\


        \bottomrule
    \end{tabular}
    
    \caption{Prompt for Description Polish.}
    \label{tab:description-polish}
\end{table*}

\endgroup
\end{CJK}

\begin{CJK}{UTF8}{gkai}
\begingroup
\begin{table*}[htp]
    \centering
    
    \begin{tabular}{p{\textwidth}}
        \toprule
        \underline{\textbf{\textsc{Prompt for Joint Translation.}}} \\
        \vspace{-2mm}
        Translate the sentence and spans from English to Chinese.\\ \\ Please follow these guidelines: \\
        1. Translate each span considering the context of the sentence.\\
        2. Ensure the number of spans after translation matches the original number of spans.\\
        3. When outputting spans, ensure only to output the translation of each span.\\

        \\
        The following is a few examples:\\ \\
        \texttt{[English]}  \\
        "sentence": "The EU rejected Germany's call for a boycott of British lamb."\\
        "spans": ["EU"]

        \texttt{[Chinese]}  \\
        "sentence": "欧盟拒绝德国呼吁抵制英国羊肉。"\\
        "spans": ["欧盟"]\\ \\

        \texttt{[English]}  \\
        "sentence": "FM involves 2 - 4.7\% of the general population."\\
        "spans": []

        \texttt{[Chinese]}  \\
        "sentence": "FM 影响了2 - 4.7\% 的普通人群。"\\
        "spans": []\\ \\

        \texttt{[English]}  \\
        "sentence": "4000 guests from home and abroad attended the opening ceremony."\\
        "spans": ["home", "abroad"]

        \texttt{[Chinese]}  \\
        "sentence": "4000名来自国内和国外的嘉宾出席了开幕式。"\\
        "spans": ["国内", "国外"]\\ \\

        Please translate the following sentence and spans: \\
        \texttt{[English]}  \\
        "sentence": "\{src\_sentence\}"\\
        "spans": [\{src\_spans\}] 
        
        \texttt{[Chinese]}  \\

        \bottomrule
    \end{tabular}
    
    \caption{Prompt for Joint Translation.}
    \label{tab:translate-prompt}
\end{table*}

\endgroup
\end{CJK}

\begin{CJK}{UTF8}{gkai}
\begingroup
\begin{table*}[htp]
    \centering
    
    \begin{tabular}{p{\textwidth}}
        \toprule
        \underline{\textbf{\textsc{Prompt for Span Rephrase.}}} \\
        \vspace{-2mm}
        Please find the Chinese span corresponding to the English span in the Chinese sentence.\\  \\
        Please follow these guidelines:\\
        1. Only find the span in the Chinese sentence that corresponds to the English span.\\
        2. Ensure that the Chinese span must be semantically consistent with the English span. 
\\ \\
        The following is an example:\\ \\

        \texttt{[English]}  \\
        "sentence": "Siemens invested 800 million US dollars to complete the electric power plant project."\\
        "spans": ["US"]

        \texttt{[Chinese]}  \\
        "sentence": "西门子投资了8亿美元完成了电力厂项目。"\\
        "spans": ["美"]\\ \\
        
Please find the corresponding span in the Chinese sentence: \\ 

\texttt{[English]}  \\
        "sentence": "\{src\_sentence\}"\\
        "spans": [\{src\_span\}]

        \texttt{[Chinese]}  \\
        "sentence": "\{tgt\_sentence\}"\\
        "spans":  \\
        

        
        \bottomrule
    \end{tabular}
    
    \caption{Prompt for Span Rephrase.}
    \label{tab:span-rephrase-prompt}
\end{table*}
\endgroup
\end{CJK}

\begingroup
\begin{table*}[htp]
    \centering
    
    \begin{tabular}{p{\textwidth}}
        \toprule
        \underline{\textbf{\textsc{Prompt for Sentence Rephrase.}}} \\
        \vspace{-2mm}
        Please translate the following sentence from English to Chinese.\\ \\
        Please follow these guidelines: \\
        1. Ensure that the translation includes the following spans: [\{tgt\_lang\_spans\}]. \\
        2. If the target sentence is semantically inconsistent with the source sentence, return "modification failure". \\ \\
        \texttt{[English]}  \\
        "sentence": "\{src\_sentence\}"\\

        \texttt{[Chinese]}  \\
        "sentence": \\

        \bottomrule
    \end{tabular}
   
    \caption{Prompt for Sentence Rephrase.}
     \label{tab:sentence-rephrase-prompt}
\end{table*}

\end{document}